\documentclass[letterpaper, 10 pt, conference]{ieeeconf}  

\IEEEoverridecommandlockouts                              
\usepackage{color}
\usepackage{float}
\usepackage{comment}
\definecolor{rosegold}{rgb}{0.72, 0.43, 0.47}
\usepackage[english]{babel}
\usepackage[utf8x]{inputenc}
\usepackage[T1]{fontenc}
\usepackage{placeins}
\usepackage{multirow}
\usepackage{graphicx}      
\usepackage{amsmath}
\usepackage{amssymb}
\usepackage{eurosym}
\usepackage{color}

\usepackage{subcaption}

\DeclareMathOperator*{\argminA}{arg\,min} 
\DeclareMathOperator*{\lSupervised}{l} 
\usepackage[linesnumbered,ruled]{algorithm2e}

\title{\LARGE \bf
Sample Efficient Interactive End-to-End Deep Learning for Self-Driving Cars with Selective Multi-Class Safe Dataset Aggregation
}

\author{Yunus Bicer$^{1}$, Ali Alizadeh$^{2}$, Nazim Kemal Ure$^{3}$, Ahmetcan Erdogan$^{4}$ and Orkun Kizilirmak$^{4}$ 
\thanks{*This work is supported by AVL Turkey and Scientific  and  Technological Research  Council  of  Turkey under the grant agreement TEYDEB 1515 / 5169901}
\thanks{$^{1}$Y. Bicer is with Faculty of Aeronautics and Astronautics, Aerospace Engineering,
        Istanbul Technical University, Turkey
        {\tt\small biceryu at itu.edu.tr}}%
\thanks{$^{2}$A. Alizadeh is with Faculty of Mechatronics Engineering, Istanbul Technical University, Turkey {\tt\small Alizadeha at itu.edu.tr}}%
\thanks{$^{3}$N.K. Ure is with Faculty of Aeronautics and Astronautics, Department of Aeronautical Engineering, Istanbul Technical University, Turkey
        {\tt\small ure at itu.edu.tr}}%
\thanks{$^{4}$A. Erdogan and O. Kizilirmak are with AVL Turkey, Istanbul, Turkey
        {\tt\small ahmetcan.erdogan, orkun.kizilirmak at avl.com}}%
}

\begin{document}

\maketitle
\thispagestyle{empty}
\pagestyle{empty}

\begin{abstract}
The objective of this paper is to develop a sample efficient end-to-end deep learning method for self-driving cars, where we attempt to increase the value of the information extracted from samples, through careful analysis obtained from each call to expert driver's policy. End-to-end imitation learning is a popular method for computing self-driving car policies. The standard approach relies on collecting pairs of inputs (camera images) and outputs (steering angle, etc.) from an expert policy and fitting a deep neural network to this data to learn the driving policy. Although this approach had some successful demonstrations in the past, learning a good policy might require a lot of samples from the expert driver, which might be resource-consuming. In this work, we develop a novel framework based on the Safe Dataset Aggregation (safe DAgger) approach, where the current learned policy is automatically segmented into different trajectory classes, and the algorithm identifies trajectory segments/classes with the weak performance at each step. Once the trajectory segments with weak performance identified, the sampling algorithm focuses on calling the expert policy only on these segments, which improves the convergence rate. The presented simulation results show that the proposed approach can yield significantly better performance compared to the standard Safe DAgger algorithm while using the same amount of samples from the expert.
\end{abstract}

\section{INTRODUCTION}

Recent years saw significant advances in self-driving car technologies, mainly due to several breakthroughs in the area of deep learning. In particular, the use of vision-based methods to generate driving policies has been of interest to a vast body of researchers, resulting in a variety of different learning and control architectures, that can be roughly classified into classical and end-to-end methods. Conventional methods approach the problem of autonomous driving in three stages; perception, path planning, and control \cite{leonard2008}. In the perception stage, feature extraction and image processing techniques such as color enhancement, edge detection, etc. are applied to image data to detect lane markings. In path planning, reference, and the current path of the car is determined based on the identified features in perception. In the control part, control actions for the vehicle such as steering, speed, etc. are calculated from reference and the current path with an appropriate control algorithm. The performance of the classical methods heavily depends on the performance of the perception stage, and this performance can be sub-optimal because of the manually defined features and rules in this stage \cite{c1}. Sequential structure of the classical methods might also lead to the non-robustness against errors, as an error in feature extraction can result in an inaccurate final decision.

On the other hand, end-to-end learning methods learn a function from the samples obtained from an expert driving policy. The learned function can generate the control inputs directly from the vision data, combining the three layers of the classical control sequence into a single step. By far, the most popular approach for representing the mapping from images to controls in end-to-end driving is using neural networks (NN). ALVINN by Pomerleau \cite{c2} is one of the initial works in this area, which uses a feedforward neural network that maps frames of the front-facing camera to steering input. Researchers from Nvidia utilized convolutional neural networks (CNN) \cite{c3} to automatize the feature extraction process and predict steering input. An FCN-LSTM architecture\cite{c4} is proposed to increase learning performance with scene segmentation. In \cite{c5}, a visual attention model used to highlight some essential regions of frames for better prediction. Although the steering input prediction in an end-to-end manner is a well-studied problem in the literature, the steering input alone is not sufficient for fully autonomous driving. In \cite{c6}, a CNN-LSTM network is proposed to predict the speed and steering inputs synchronously.

Pure end-to-end learning policies are limited to the demonstrated performance, and although the training and validation loss on the data collected from the expert might be low, errors accumulated from the execution of the learned driving policy might lead to poor performance in the long run. This performance loss is partly because the learned driving policy is likely to observe states that do not belong to the distribution of the original expert demonstration data. DAgger \cite{c7} algorithm addresses this issue by iteratively collecting training data from both expert and trained policies. The main idea behind DAgger is to actively obtain more samples from the expert to improve the learned policy. Even though DAgger achieves better driving performance, it might end up obtaining a lot of samples from the expert, which can be time and resource-consuming in many real-life scenarios. SafeDAgger \cite{c8} algorithm, an extension of DAgger, attempts to minimize the number of calls to the expert by predicting the unsafe trajectories of the learned driving policy and only calls the expert on such cases. Another extension of DAgger, EnsembleDAgger \cite{c9}, predicts the variance of the decisions by using multiple models and takes it as additional safety criteria like SafeDAgger.

In this paper, we propose a novel framework which is sample-efficient compared to the SafeDAgger algorithm (state-of-the-art data aggregation method), named Selective SafeDAgger. The proposed algorithm classifies the trajectories executed by the learned policy to safe and multiple classes of unsafe segments. After the prediction, the model focuses on obtaining the expert policy samples primarily from the identified unsafe segment classes. Our main contribution is an imitation learning algorithm that collects the most promising samples from the expert policy, which enables outperforming the SafeDAgger method while limited to the same number of calls to the expert.

This paper is organized as follows. Section II provides the details of the methodology. The experimental setup is provided in section III, followed by a discussion about results in section IV and conclusions in section V.

\section{METHODOLOGY}

In this section, driving policies, the architecture of the network, and devised algorithm are explained in detail.

\subsection{Driving Policies}

We begin with giving definitions of the used terms to explain driving policies in detail.  

A set of states $S$ for the car in this paper is an environment model, and $s \in S$ is one of the states for the car in that environment. Observation of the state $s$ is defined as $\phi(s) \in \Phi(S)$ where $\Phi(S)$ is the observation set for all states. $a(s)\in A(S)$ will be driving action at observation $\phi(s)$ where $A(S)$ is the set of all possible actions.  

A set of driving policies $\Pi$ is defined as in Eq. (\ref{eq:drivingPloicy}).\begin{equation} \label{eq:drivingPloicy} 
\Pi: \Phi(S) \rightarrow A(S) 
\end{equation}where $\Pi$ is a mapping from state observations $\phi(s)$ to driving actions $a(s)$ such as steering, throttle, brake, etc. 

Two distinct driving policies are defined throughout the paper. The first one is an expert policy $\pi^*\in\Pi$ that drives the car with a reasonable performance that we want to imitate. An expert policy in an autonomous driving scenario is usually chosen as actions of a human driver. Variants of DAgger algorithms, however, have mislabeling problem in case of the human driver, since drivers do not have feedback feelings from their actions and they can give incorrect reactions to the given states. To overcome the mislabeling problem, we have used a rule-based controller which contains speed and steering controllers, as an expert policy in this paper. The second one is a primary policy $\pi_0\in\Pi$ that is trained to drive a car. This policy is a sub-optimal policy according to the expert policy since it is trained on a subset of observation set $\Phi(S)$.

Training a primary policy to mimic an expert policy is called imitation learning or learning by demonstration. One of the most common methods for imitation learning is based on supervised learning techniques. The loss function for the supervised learning is defined as in Eq. (\ref{eq:loss}) \cite{c8}. \begin{equation} 
\label{eq:loss}
l_{supervised}(\pi,\pi^{*},D_0) = \frac{1}{N}\sum_{i=1}^{N} ||\pi(\phi(s_i)) - \pi^{*}(\phi(s_i))||^2 
\end{equation}where $l_{supervised} $ refers to $l^2$-Norm between trained and expert policy actions.     

A primary policy as in Eq. (\ref{eq:primary}) is defined as a policy that minimizes the loss function as follows.\begin{equation} 
    \label{eq:primary}
        \pi_{0} = \argminA_{\pi} l_{supervised}(\pi, \pi^{*}, D_{0})
\end{equation}

Minimization of the loss function can be challenging since it is known that the relation between image frames and driving actions is highly nonlinear. So, we have used a deep neural network architecture to find an optimal solution for the primary policy.

\subsection{Network Architecture}
The earlier works in end-to-end learning for self-driving cars focus on computing only the steering angle from a single image or a sequence of images. The longitudinal control component is required to reach a higher level of autonomy in the end-to-end framework. In this work, we utilize the multi-task model proposed in \cite{c6} as our baseline, which is capable of generating both longitudinal and lateral control inputs for the car. Besides, we utilize a speed controller rather than the classical throttle/brake commands for the longitudinal control. The steering action is predicted from the raw image inputs taken from the cameras located in front of the vehicle through convolution layers, and the speed is predicted from a sequence of speed profiles through a Long-Short Term Memory (LSTM) layer. There exists a single-direction coupling between the longitudinal controller (speed controller) and the lateral steering actions. In particular, the speed of the vehicle has a significant impact on the prediction model, since entering a turn with low speed represents different dynamics for the lateral controller when compared to a high-speed maneuver. Moreover, the straight trajectory dominates the whole other trajectory types (e.g., turn left, turn right); therefore, the trained network will be biased toward the straight path. To recover from this issue, we decided to define various trajectory types including all major maneuvers such as straight, turn left, turn right and low and high-speed scenarios, by which the devised model will learn the other less-occurring maneuvers. 

\begin{figure}[]
	\centering
	\includegraphics[width=0.8\columnwidth,trim={0 0.46cm 0 0.33cm},clip]{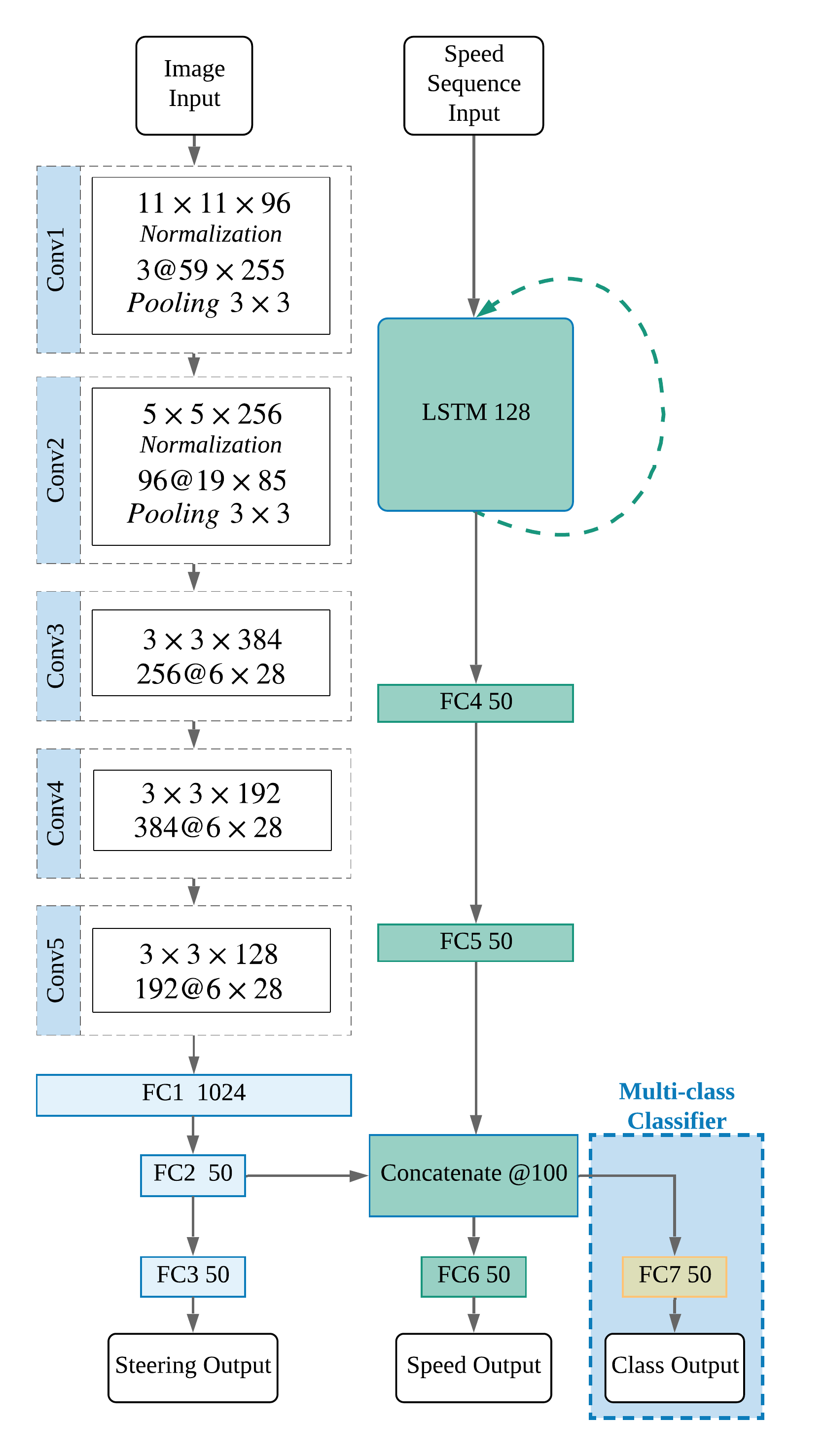}
	\caption{Sample-efficient Selective SafeDAgger model}
	\label{fig:Architecture}
\end{figure}

The model architecture is shown in Fig \ref{fig:Architecture}. It takes the current observation and the past speed profile and returns steering action, speed action, and the class of the trajectory segment. The multi-task network predicts the steering angle through a visual encoder using a stack of convolution and fully-connected layers.  In the first two convolution layers (Conv1 and Conv2), large kernel size is adopted to better capture the environment features, which is suitable for the front-view camera. Inputs and kernels of the each convolution layer is denoted by "$\# channels @ input\,height \times input\,width $" and "$kernel\,height \times kernel\,width \times \# channels$" and each fully connected layer is denoted by "$FC - \text{size of neurons}$". The speed and trajectory class are predicted through a concatenation of visual encoder and feedback speed features. The speed features are extracted by an LSTM layer followed by fully-connected layers. ReLU (Rectified Linear Unit) is used as the activation function for all layers. Mean absolute error is the loss function for both speed and steering angle predictions as regression problems. On the other hand, the cross-entropy applies to the trajectory classifier as a classification problem.

The multi-class classifier highlighted in Fig. \ref{fig:Architecture} extends the safeDAgger method to a novel algorithm devised in this paper. The trajectory classes are defined as follows:\begin{equation} \label{eq:multi-class}
c_{s}(\pi,\phi(s)) =     \begin{cases}
       1,&\text{Safe  Trajectories} \\
       2,& \text{Unsafe Low-Speed Left(LL)} \\
       3,&\text{Unsafe High-Speed  Left(HL)} \\
       4,&\text{Unsafe Low-Speed  Right(LR)} \\
       5,&\text{Unsafe High-Speed  Right(HR)} \\
       6,&\text{Unsafe Low-Speed Straight(LS)} \\
       7,&\text{Unsafe High-Speed Straight(HS)} \\
      \end{cases}
  \end{equation}Low and high speeds with combinations of left, straight and right turn cover almost all unsafe trajectories. Same combinations also applicable for safe trajectories but since it is not needed to call expert policy in safe trajectories, we define only one class for the safe trajectories.

The multi-class classifier takes the partial observation of the state $\phi(s)$ which contains the visual perception and the past speed profile and returns a label indicating in which part of the trajectory the policy will likely to deviate from the expert policy $\pi^{*}$.

The labels for training the model was generated through one-hot-encoding method, defined by sequential decisions; first, it was decided whether the policy is safe by measuring its distance from the expert policy through $l^2$-Norm metric using Eq. (\ref{eq:l2_norm_metric}). \begin{equation}
\label{eq:l2_norm_metric}
c_{s}(\pi, \phi(s)) = \begin{cases}
0, & ||\pi(\phi(s)) -\pi^{*}(\phi(s))|| > \tau_{safe} \\  
1,&\text{otherwise}\end{cases}
\end{equation}where $\tau_{safe}$ is a predefined threshold and can be chosen arbitrarily. Furthermore, to distinguish between low-speed and high-speed turn trajectories, steering threshold $\tau_{turn}$, speed thresholds for turn maneuver $\tau_{speed, turn}$ and straight trajectory $\tau_{speed, straight}$ are defined heuristically based on the response of the car dynamics in these trajectories. The threshold values for this work is depicted in Table \ref{table:thresholds}.
\vspace{-3mm}

\begin{table}[h]
{\setlength{\tabcolsep}{14pt}
\caption{Threshold Values in Labeling Process}
\begin{center}
\vspace{-3mm}
\begin{tabular}{cc}
\hline\hline
Parameter & Threshold value \\
\hline
$\tau_{safe}$ & $0.5$ \\
$\tau_{turn}$ & $0.25^{\circ}$\\
$\tau_{speed, turn}$ & $10 \, \, m/s$ \\
$\tau_{speed, straight}$ & $13.75 \, \, m/s$ \\
\hline
\end{tabular}
\end{center}
\label{table:thresholds}}
\vspace{-4mm}
\end{table}\noindent where $\tau_{safe}$ as $0.5$ yields $0.25^{\circ}$ for the steering angle and $1 \,  m/s$ for the speed difference between the network prediction and expert policy output.

\subsection{Selective SafeDAgger Algorithm}

\begin{algorithm}

\SetKwInOut{Input}{Input}
\SetKwInOut{Output}{Output}
Collect $D_{0}$ using $\pi^{*}$ \label{D_0} \\
$\pi_{0} = \argminA_{\pi} l_{supervised}(\pi, \pi^{*}, D_{0})$ \\
\For{i = 1:N}{ 
    \label{begin}
    \textcolor{blue}{{$c_{i} \leftarrow $ Define unsafe classes over $D_{0}$} \\      $D’ \leftarrow [\,]$  \\
    \While{$k \le T$}{\label{begin2}
        $\phi_k \leftarrow  \phi(s)$\\
        $c_{\phi_k}  \leftarrow$ classifier output of $\pi_i(\phi_k)$ \\
        \eIf{$c_{\phi_k}  \in c_i$}{
            use $ \pi^*(\phi_k)$\\
            $D^{'} \leftarrow [\phi_k]$\\
            $k = k+1$
        }
        {
            use $\pi_i(\phi_k)$\\
        }
    }\label{end2}}
    $D_{i} = D_{i-1} \cup D^{'}$ \\
    $\pi_{i+1} = \argminA_\pi \,\lSupervised_{supervised}(\pi, \pi^{*}, D_{i})$ \\
    \label{end}
}
\textbf{return} best $\pi_i$ over validation set
\caption{Selective SafeDAgger: \quad \small \textcolor{blue}{Blue} fonts distinguishes the difference between Selective SafeDAgger and SafeDAgger}
\label{Algorithm1}
\end{algorithm}

Algorithm \ref{Algorithm1} describes the proposed method in detail, which takes the expert policy $\pi^{*}$ as an input and gives $\pi_i$ as an output. The primary dataset $D_0$ is collected by using $\pi^*$, which is then utilized in training a primary policy $\pi_0$ by a supervised learning method. Having the $\pi_0$ at hand,  $c_i$, the unsafe classes of $D_0$ for the trained policy $\pi_i$ are determined. An observation $\phi_k $ taken from environment $\phi(s)$ is evaluated by $\pi_i$ to find its class $c_{\phi_k}$. If  $c_{\phi_k}$ is an element of $c_i$, $\pi^{*}$ takes over the control of the car and  $\phi_k $ is appended to $D^{'}$. Otherwise, $\pi_i$ continues to command the car until it encounters an unsafe class. As depicted in lines \ref{begin2}-\ref{end2}, the algorithm continues to append data to $D^{'}$ for T number of iterations. The appended dataset $D^{'}$ is aggregated into  $D_{i-1}$ to create $D_i$ and $\pi_{i+1}$ is trained on $D_i$. This loop is repeated for N times, as shown in lines \ref{begin}-\ref{end}. In the end, the algorithm returns the best $\pi_i$ over the validation set.

\section{EXPERIMENTS}

\subsection{System Setup}

\subsubsection{Simulator}

AirSim used in this work is an Unreal Engine Plugin based simulator for drones and cars established by Microsoft to create a platform for AI studies to develop and test new deep learning, computer vision and reinforcement learning algorithms for autonomous vehicles with photo-realistic graphs in simulations \cite{c10}. It has built-in APIs for interfacing with Python coding language. Furthermore, the engine editor creates custom environments or scenarios.

\begin{figure}[h]
    \centering
    \includegraphics[width=0.9\columnwidth,trim={0 1.3cm 0 2.3cm},clip]{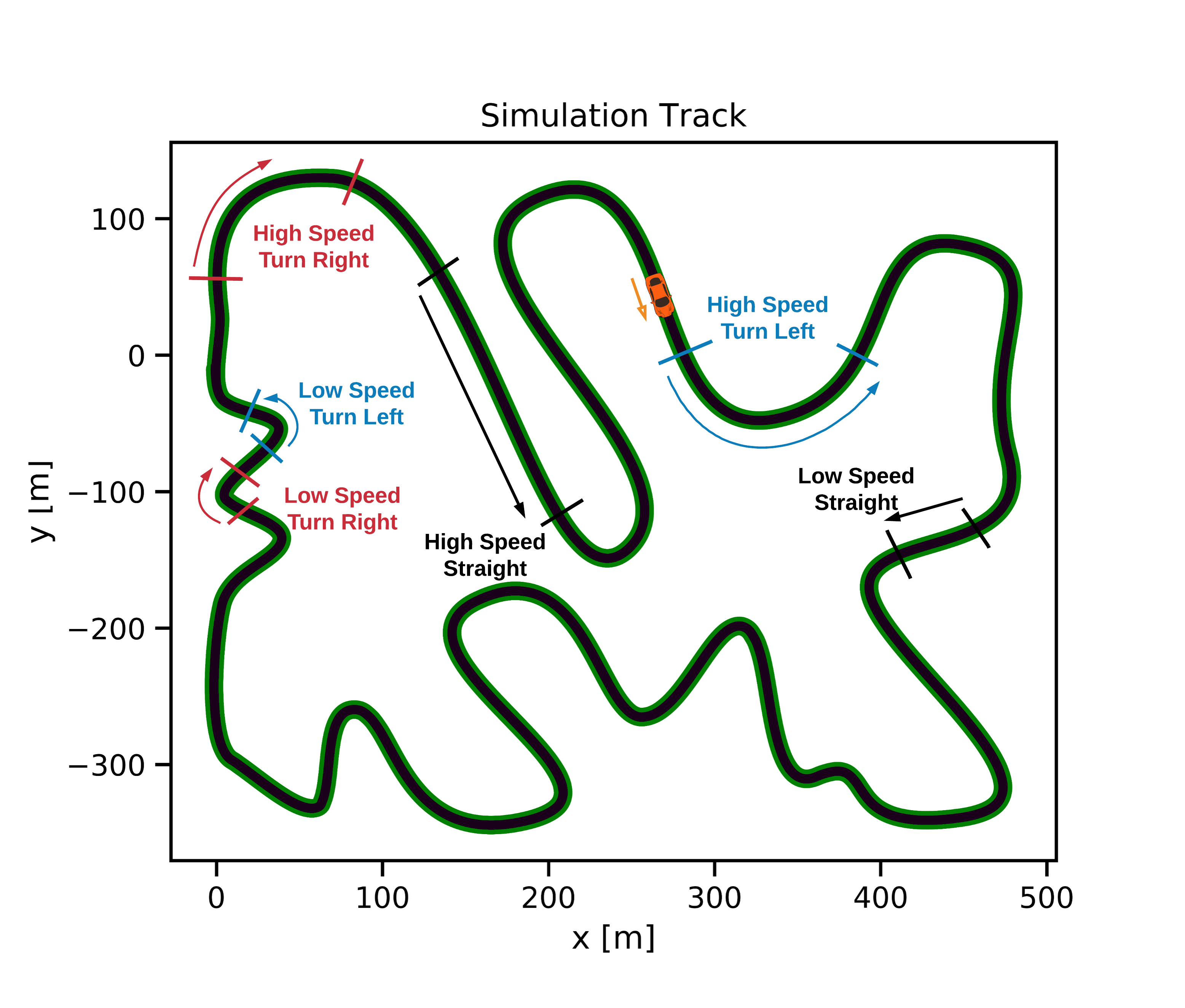}
    \caption{Train set track}
    \label{fig:track_visualization}

\end{figure}

The road track for the training process of the algorithm is devised in a way to capture all defined scenarios in this work. The geometry of the custom created training track is shown in Fig. \ref{fig:track_visualization}, in which all the trajectory classes are illustrated.  

Representative power of the training set can be increased by collecting data from unseen observations. With that reason, two additional cameras were added to the front-facing camera with an angle of $\gamma$ to imitate turning-like maneuvers \cite{c3}. Airsim APIs provide ground truth labels for the front-facing camera frames, but ground truth labels for the left and right cameras should be adjusted with a threshold as in Eq. (\ref{eq:3cam_eqn}).

\begin{equation} \label{eq:3cam_eqn}
    \begin{bmatrix}
    L_{l} \\
    L_{r}
    \end{bmatrix} =
        \begin{bmatrix}
    L_{c_{steering}} +\gamma &L_{c_{speed}} - p_{speed} \\
    L_{c_{steering}} - \gamma & L_{c_{speed}} -  p_{speed}
    \end{bmatrix}
\end{equation}where $L_{l}$, $L_{r}$, $L_{c_{steering}}$ and $L_{c_{speed}} $ refer to the ground truth for the left and right cameras, center camera steering and speed actions respectively. In the turning case, the ground truth speed of the vehicle is adjusted by a parameter $p_{speed}$ which is chosen as $4$ $m/s$ heuristically.



\subsubsection{Data Preprocessing}
A couple of techniques were utilized in the preprocessing phase. The input raw image was down-sampled to the size of 144×256×3 (RGB) and a Region of Interest (ROI) defined with the size of 59×255 to cover almost the entire road and ignore the features above the horizon, which reduces the computational cost. Moreover, to improve the convergence rate and robustness of the neural network model, the processed image was normalized to the range [0,1] and augmented by randomly changing the brightness of the image with a scale of 0.4. The normalization was done by dividing all image pixels by 255.


\subsubsection{Expert Policy }

\begin{figure}[h]
    \centering
    \includegraphics[width=0.75\columnwidth]{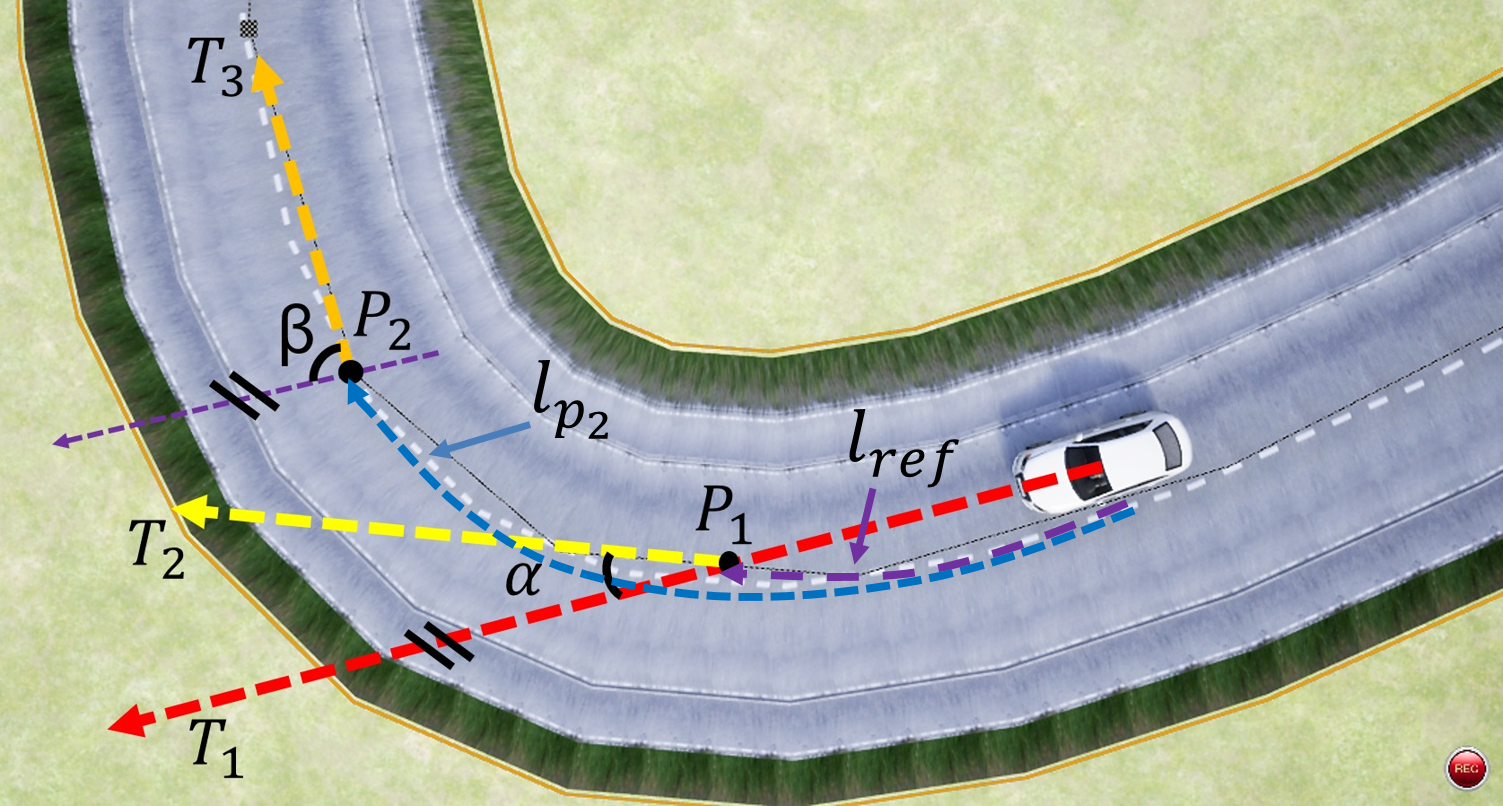}
    \caption{Expert policy}
    \label{fig:expert}
\end{figure}

To automatize the data collection part of the algorithm, a rule-based expert policy is defined as shown in Fig. \ref{fig:expert}. 

For the steering action, $T_1$ is a tangent line to the road spline at the position of the car and $P_{1}$ is a point on road spline with $l_{ref}$ distance along spline from that positions. Tangent line at $P_{1}$ according to road spline is $T_2$. The angle between $T_1$ and $T_2$ which is $\alpha$ will be expert steering action as depicted in Eq. (\ref{eq:SteeringCommand}).\begin{equation}
    \label{eq:SteeringCommand}
     a_{steering}= \alpha = \arccos{\left(\frac{ T_1\cdot T_2}{\left\lVert T_1\right\rVert \left\lVert T_2\right\rVert}\right)}
\end{equation}
For the speed action, $P_2$ is a point on the road spline with a distance $l_{P_2}$ from the position of the car along the road spline as depicted in Eq. (\ref{eq:RefDist}). \begin{equation}
    \label{eq:RefDist}
    l_{P_2} = l_{ref} V_{current} k_{steering}
\end{equation}

\noindent where $V_{current}$ is current speed and $k_{steering}$ is a fine tuned constant. Tangent line at $P_2$ according to the road spline is $T_3$. 

\begin{figure*}
    \centering
    \includegraphics[width=0.8\linewidth,trim={0 0.52cm 0 0.5cm},clip]{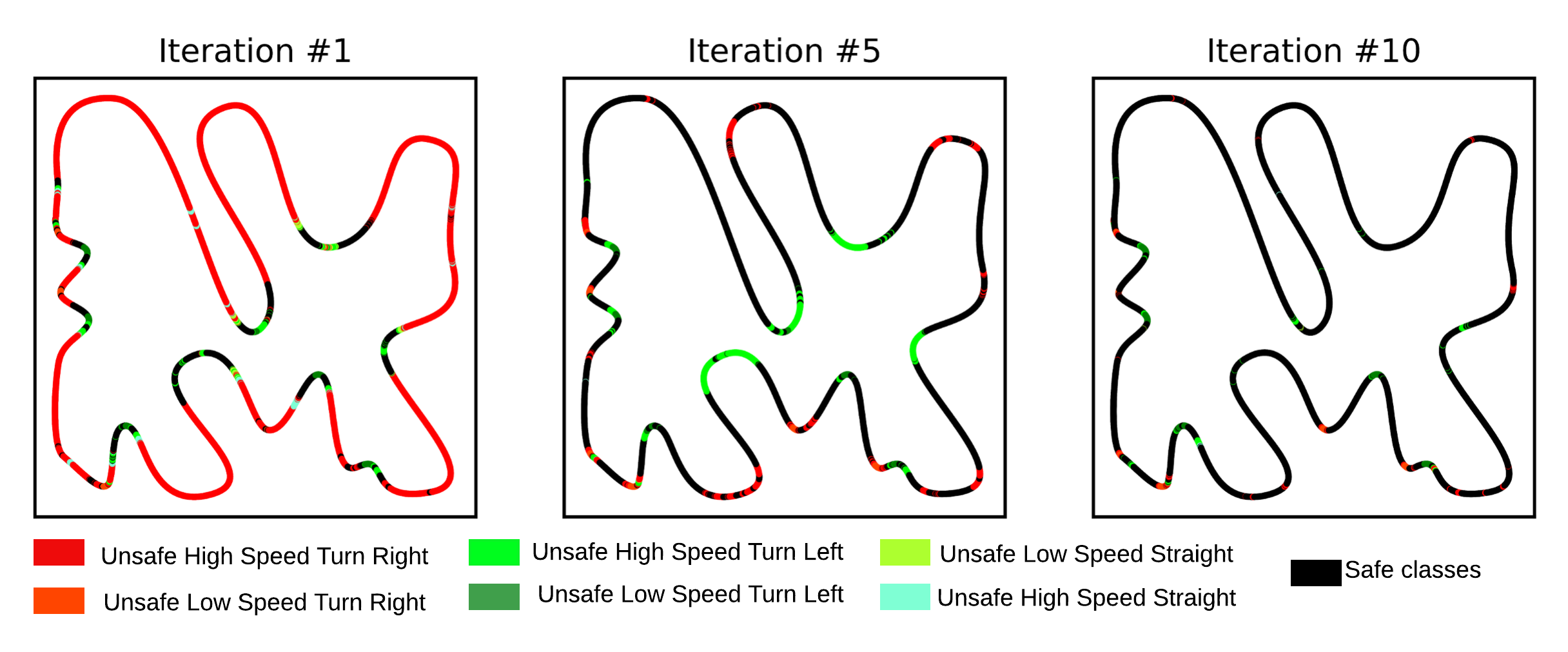}
    \caption{Convergence rate of the proposed model; It shows the improvement of the model as the number of dataset aggregation iterations increases.}
    \label{fig:3-iteration_convergence}
    \vspace{-4mm}
\end{figure*}

Expert speed action is defined by Eq. (\ref{eq:SpeedCommand}). \begin{equation}
    \label{eq:SpeedCommand}
    a_{speed} = V_{cruise}-\beta k_{speed}
\end{equation}

\noindent where $V_{cruise}$ is a pre-defined cruise speed, $k_{speed}$ is a fine tuned gain and $\beta$ is an angle between $T_1$ and $T_3$.  

For our implementation, the parameters are chosen as $l_{ref} = 1$ m, $k_{steering} = 5 $,  $V_{cruise}=13.8$ m/s and $k_{speed}=10$.

\subsection{Training}

For the training of the primary policy $\pi_0$, dataset $D_0$, which contains 2800 image data were collected by using expert policy $\pi^*$. Nesterov Adam Optimizer (Nadam) was used as an optimizer for the training of the network with the initial learning rate of $10^{-5}$ and moment of 0.99. The Training continued for ten epochs with the batch size of 32.

Trained primary policy $\pi_0$ is tested on the pre-collected dataset to classify trajectories and calculate the $l^2$-Norm of each sample in the dataset. The weakness of the network over trajectory segments is determined by a coefficient of weakness, which is defined as in Eq. (\ref{eq:class_weak}). \begin{equation}
\label{eq:class_weak}
c_{i} = \dfrac{N_{L2_i}}{N_{i}} \times \mu_{L2,i}
\end{equation}

\noindent where $\mu$, $\sigma$ are mean and standard deviations for the $l^2$-Norm of class$_i$. $N_{L2_i}$ is the total number of samples in class$_i$ that $l^2$-Norm of samples fall in the region of one $\sigma$ away from the mean $\mu$. $N_{i}$ is the total number of samples in class$_i$.

Once the weakness coefficients are calculated, trajectory classes are sorted according to their weakness coefficients, and the two of the most dominant unsafe classes will be chosen for data aggregation as shown in Table \ref{table:coeff}. Additionally, the classes with the mean $l^2$-Norm lower than 1, will be selected as allowable classes.

As depicted in Table \ref{table:coeff}, the weakness coefficients for the class $LS$ and $HS$ are quite low and never chosen as weak classes. The initial dataset for the training is biased toward $LS$, and $HS$ classes and $l^2$-Norms in those classes are low, which lead to low weakness coefficients. Moreover, training track does not have many samples from class $LR$ so that weakness coefficients for the class $LR$ is also low. 

\begin{table}[h]
{\setlength{\tabcolsep}{7pt}
\caption{Coefficient of weakness for each class}
\begin{center}
\vspace{-3mm}
\begin{tabular}{ccccccc}
\hline\hline
\# Iter. &  $LL$ &  $HL$  &  $LR$  &  $HR$  &  $LS$ & $HS$ \\ \hline
1 & 0.004 & \textbf{0.321} & 0.019 & \textbf{0.694} & 0.002 & 0.010 \\ 
2 & \textbf{0.505} & 0.122 & 0.037 & \textbf{0.278} & 0.001 & 0.023 \\ 
3 & \textbf{0.635} & 0.264 & 0.028 & \textbf{0.607} & 0.001 & 0.062 \\ 
4 & \textbf{0.751} & 0.515 & 0.046 & \textbf{0.646} & 0.001 & 0.010 \\ 
5 & 0.018 & \textbf{0.678} & 0.034 & \textbf{0.755} & 0.001 & 0.010 \\ 
6 & 0.009 & \textbf{0.752} & 0.039 & \textbf{0.849} & 0.000 & 0.006 \\ 
7 & 0.717 & \textbf{0.790} & 0.038 & \textbf{0.780} & 0.001 & 0.004 \\ 
8 & 0.028 & \textbf{0.787} & 0.017 & \textbf{0.794} & 0.001 & 0.006 \\ 
9 & \textbf{0.670} & 0.634 & 0.011 & \textbf{0.713} & 0.001 & 0.005 \\ 
10& 0.012 & \textbf{0.768} & 0.020 & \textbf{0.809} & 0.001 & 0.003 \\ 
\hline
\end{tabular}
\end{center}
\label{table:coeff}}
\vspace{-3mm}
\end{table}

After determination of the weak and allowable classes, the data aggregation phase begins. In this phase, policy $\pi_i$ drives the car to collect 10 batches of data in dominant classes. If policy faces with dominant classes, the expert policy takes control of the vehicle and samples are taken in that time labeled and reserved for aggregation. If policy $\pi_i$ faces with allowable classes which are unsafe, it continues to drive the car. For all the other unsafe classes, the expert policy takes control of the vehicle with the query limit of 10 batch-size. When the number of query reaches the limit, data aggregation freezes and training starts with the new aggregated dataset $D_i$.

After the training, $\pi_i$ becomes $\pi_{i+1}$, and determination of dominant weak classes on the pre-collected data is repeated for collecting relevant data. This process will be repeated for 10 iterations. As shown in Fig. \ref{fig:3-iteration_convergence}, in the dataset aggregation iteration number 1, a significant fraction of dataset is unsafe, and as it proceeds to recover from the most problematic cases, the model error converges. The progress of this process can be seen from iteration number 1 to 10.

\section{RESULTS}

\begin{figure}
	\centering
	\vspace{-2mm}
	\begin{subfigure}[b]{0.5\textwidth}
		\centering
		\includegraphics[width=0.80\linewidth,trim={0.2cm 0 1.2cm 1.2cm},clip]{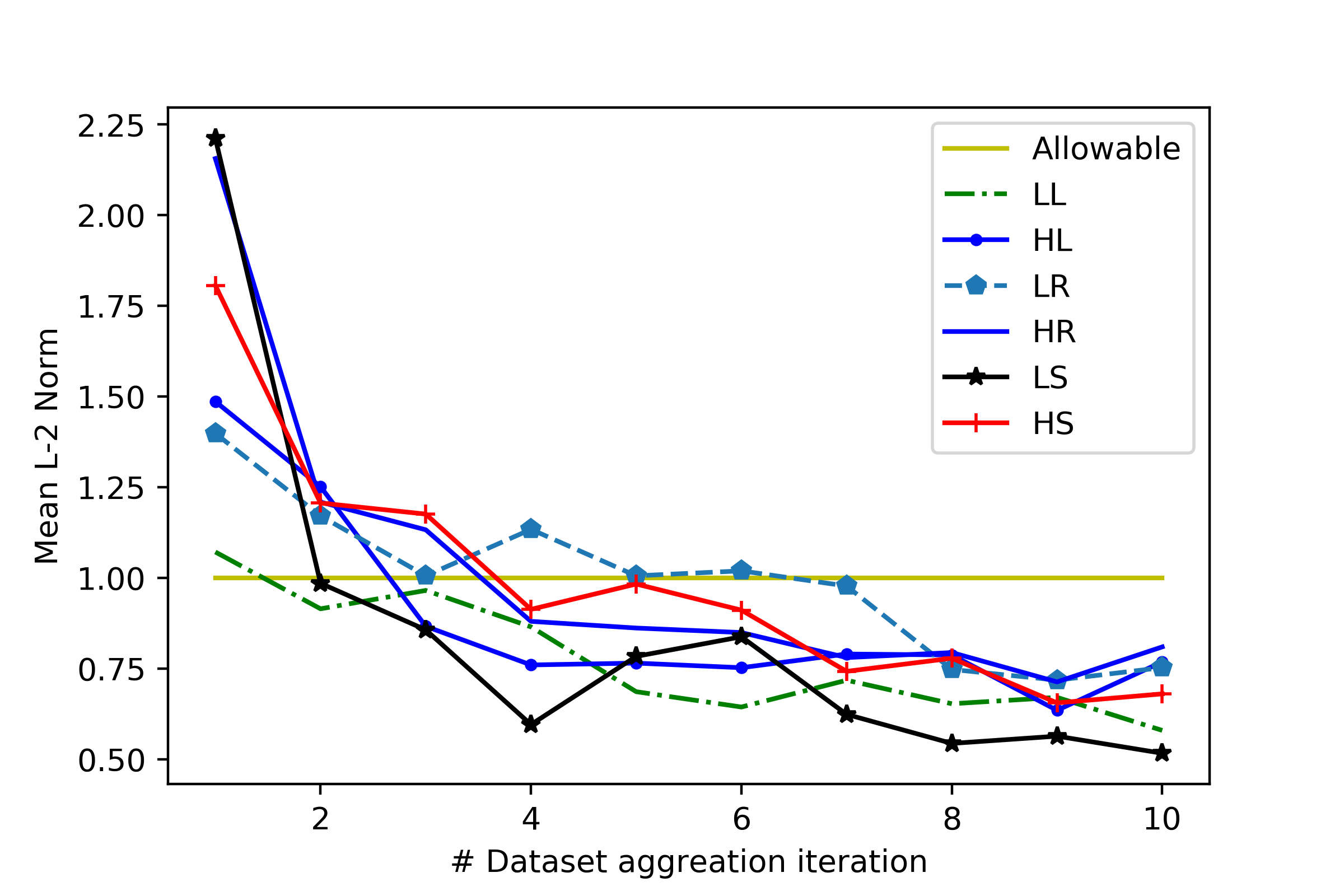}
		\caption{}
		\label{fig:Norm} 
	\end{subfigure}
	
	\begin{subfigure}[b]{0.5\textwidth}
		\centering
		\includegraphics[width=0.80\linewidth,trim={0.2cm 0 1.2cm 1cm},clip]{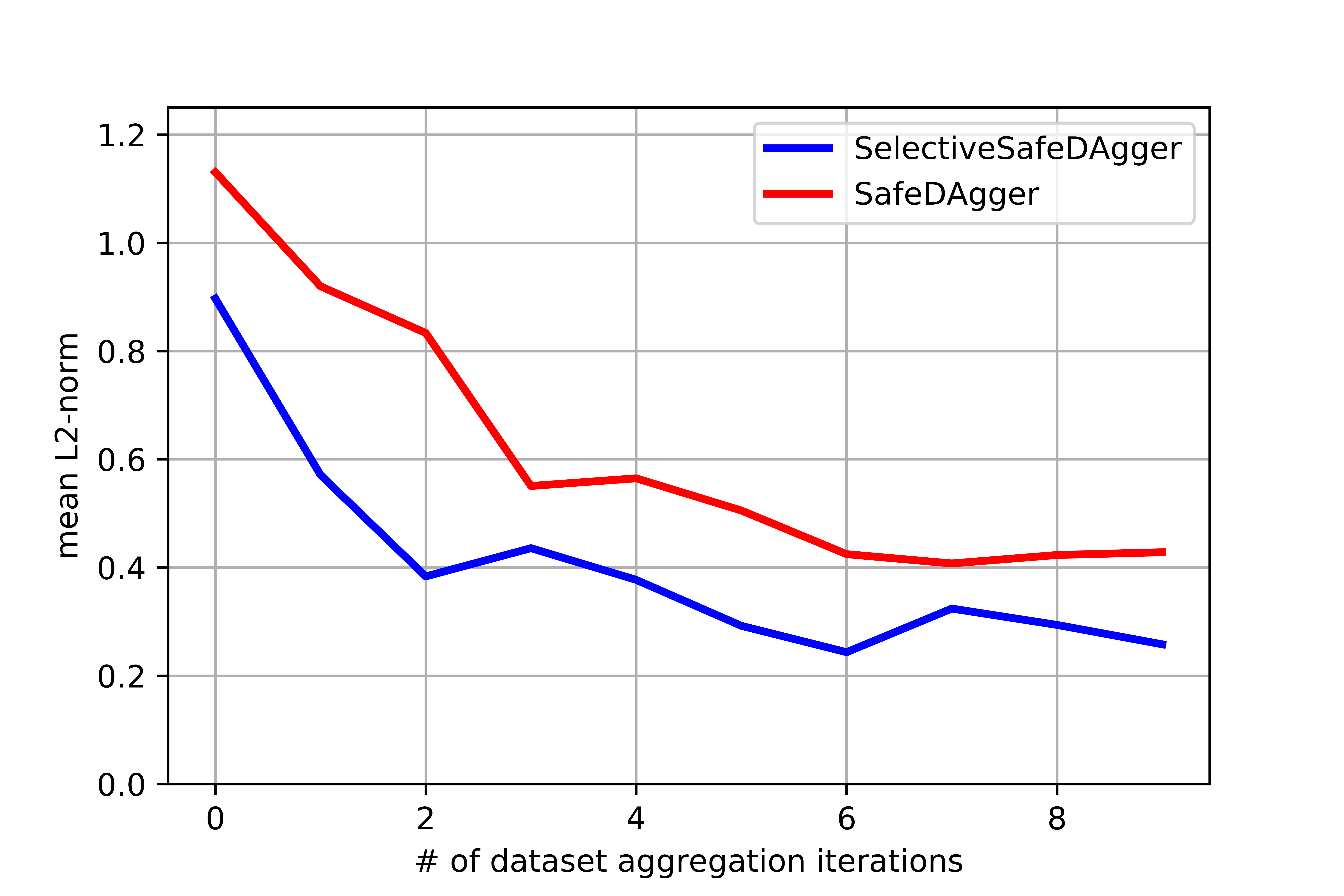}
		\caption{}
		\label{fig:Norm2}
	\end{subfigure}
	
	\caption[Two numerical solutions]{(a) Performance of the Selective SafeDAgger algorithm for all classes at each aggregation iteration. (b) $l^2$-Norm of prediction and ground truth over 10000 samples at each iteration.}
\end{figure}

In Fig. \ref{fig:Norm}, we present the performance of the Selective SafeDAgger with using metric of $l^2$-Norm in each class during the training process. For the first iteration, $HR$ and $HL$ are chosen as weak classes and data for new dataset comes from those classes by querying expert policy. It is seen that in the second iteration, $l^2$-Norms drops for all classes by using aggregated dataset. Notice that the performance of the policy for the other classes is also increased without querying expert policy for those classes which are not the case for the SafeDAgger. Sequential decision making is the main idea behind this behavior. In SafeDAgger, when policy shifts from nominal conditions, the expert policy is called, and the new dataset is collected until the safety criterion is met, which leads to an unnecessary query of the expert policy. On the other hand, Selective SafeDAgger tries to solve the problem from the beginning by finding problematic classes. Besides, after the seventh iteration, the norm of all classes drops below the allowable threshold, which means that resultant dataset covers almost all trajectory classes as seen in Fig. \ref{fig:3-iteration_convergence}.

The trained model is tested at each iteration by taking 10000 samples from the environment and mean $l^2$-Norms are calculated, accordingly. Fig. \ref{fig:Norm2} shows that selective SafeDAgger method has better performance in all iterations than the SafeDAgger method even though both ways have the same amount of query to the expert as depicted in Table \ref{table:query}.


\vspace{-2mm}
\begin{table}[h]
	{\setlength{\tabcolsep}{5pt}
		\caption{Query to expert }
		\begin{center}
			\vspace{-3mm}
			\begin{tabular}{cccccccc}
				\hline\hline
				\multirow{2}{*}{} &  \multicolumn{6}{c}{Selective SafeDAgger}  & SafeDAgger \\
				\cline{2-7}   &  $LL$ &  $HL$  &  $LR$  &  $HR$  &  $LS$ & $HS$ &  unsafe \\
				\hline
				Iteration 1     &  0 & 127  &  38 &  155 &  0 &  0 & 320\\ 
				Iteration 2     &  0 & 44  & 0  & 228  &  0 & 48  &  320\\ 
				Iteration 3     &  19 &  63 & 0  & 238  & 0  & 0  & 320\\ 
				Iteration 4     & 27  & 12  & 0  & 281  & 0  & 0  & 320\\ 
				Iteration 5     &  0 & 165  & 0  & 155  & 0  & 0  & 320\\ 
				Iteration 6     & 31  & 189  & 0  & 100  & 0  & 0  & 320\\ 
				Iteration 7     & 0  & 93  & 0  & 227  & 0  & 0  & 320\\ \
				Iteration 8     &  2 & 162  & 0   & 156  & 2  & 5  & 320\\ 
				Iteration 9     &  83 & 0  & 0  & 237  & 0  & 0  & 320\\ 
				Iteration 10    &  0 & 205  &  0 & 115  & 0  & 0  & 320\\ 
				\cline{2-7}
				Total         &  \multicolumn{6}{c}{3200} & 3200\\ 
				\hline
			\end{tabular}
		\end{center}
		\label{table:query}}
	\vspace{-6mm}
\end{table}

\begin{figure}[h]
	\centering
	\includegraphics[width=0.9\columnwidth]{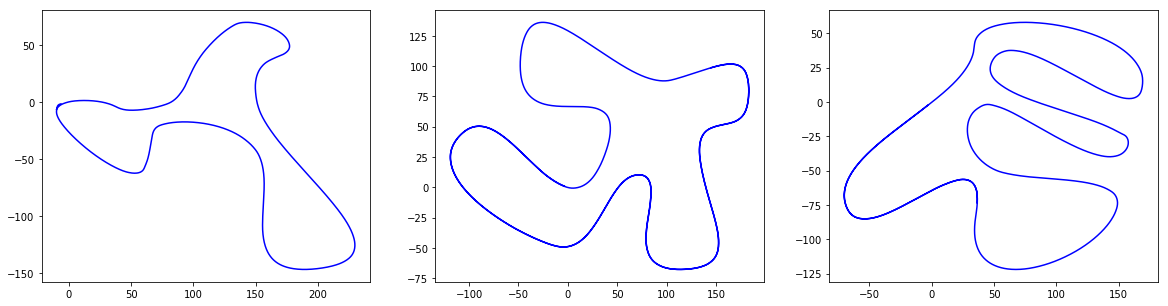}
	\caption{Geometry of test tracks.}
	\label{fig:3test}
\end{figure}

Three unseen test tracks were devised to evaluate the generalization performance of the proposed method, where their layouts are illustrated in Fig. \ref{fig:3test}. The generalization performance of the Selective SafeDAgger is depicted in Table \ref{table}, which shows its superiority over SafeDAgger method. The selectivity of the proposed algorithm will define the unsafe cases that dominate all other classes, which results in faster convergence of the model error compared to different dataset aggregation methods. 

\begin{table}[H]
	{\setlength{\tabcolsep}{10pt}
		\caption{Mean $l^2$-Norm on Unseen Test Track}
		\begin{center}
			\vspace{-3mm}
			\begin{tabular}{ccc}
				\hline\hline
				& Selective SafeDAgger & SafeDAgger \\
				\hline
				1. Test Track & \textbf{0.4794} & 0.5518\\
				2. Test Track & \textbf{0.3295} & 0.4986\\
				3. Test Track & \textbf{0.3254} & 0.3632\\
				\hline
			\end{tabular}
		\end{center}
		\label{table}}
	\vspace{-2mm}
\end{table}

\section{CONCLUSIONS}

In this work, we implemented a Selective SafeDAgger algorithm which is sample-efficient in the selection of dataset aggregation. The proposed algorithm evaluates the performance of the trained policy and determines the weakness of the policy over different trajectory classes and recovers the policy from those specific trajectory classes. Our method outperforms the SafeDAgger algorithms in term of sample-efficiency and convergence rate. Next, we aim to cluster the trajectories with unsupervised neural network techniques to have a better realization of the road trajectories. 

\addtolength{\textheight}{-0cm}   





\section*{ACKNOWLEDGMENT}

This  work  is  supported  by \textbf{Scientific  and  Technological Research  Council  of  Turkey} (Turkish:\textbf{TÜBİTAK}) under the grant agreement \textbf{TEYDEB 1515 / 5169901}.


\end{document}